\documentclass[10pt,letterpaper]{article}

\usepackage{iccm}
\usepackage{amssymb, amsfonts, amsmath}
\usepackage{graphicx}

\iccmfinalcopy 

\usepackage{pslatex}
\usepackage{apacite}
\usepackage{float} 

\usepackage{xurl}

\title{Learning a Vector-Symbolic Model for Socio-Cultural Tasks}
 
\author{{\large \bf Meera Ray (mfr5832@psu.edu)} \\
  Department of Computer Science and Engineering, The Pennsylvania State University \\
  W209 Westgate Building, University Park, PA 16802. USA
  \AND {\large \bf Swapnika Dulam (szd5775@psu.edu)} \\
   Department of Computer Science and Engineering, The Pennsylvania State University  \\
 W209 Westgate Building, University Park, PA 16802. USA
  \AND {\large \bf Christopher L. Dancy (cdancy@psu.edu)} \\
 Department of Industrial and Manufacturing Engineering, The Pennsylvania State University \\
  M310 Leonhard Building, University Park, PA 16802. USA
  }

\begin{document}

\maketitle

\begin{abstract}
How can we better represent the impact of sociocultural structures on decision making in computational cognitive models? Modeling this impact requires traversing multiple levels of semantic representation, however it is not immediately clear to a modeler which levels of representation are most salient to a given situation. Though large language models and cognitively grounded corpus models can represent broad semantic associations through co-occurences, the role of self representations in memory should be accounted for to determine how cultural associations shape decision making. We propose a declarative memory system to be used in the ACT-R cognitive architecture that represents semantic associations at multiple levels via a vector-symbolic autoencoder. We use a simple HRR operation to encode episodic memories differently from semantic memory vectors extracted from text to produce a final chunk activation for a memory request. We use ACT-R cognitive models of a racially contextualized implicit association test (IAT) to test this new declarative memory system.

\end{abstract}

\section{Introduction}
In this paper, we aim to build a vector-symbolic extension to the ACT-R model for tasks that require social band representation. ACT-R is a hybrid architecture that uses symbolic representations of knowledge, accessed via sub-symbolic numeric values established through rigorous model fitting to human data. Additionally, ACT-R is open-source, making it easy to modify required modules within the architecture, and it has an active development community.


\section{Related Work}
\subsection{Declarative Memory in ACT-R} 

The declarative memory system in canonical ACT-R encompasses episodic and semantic memory 
systems (though does not explicitly represent the two types of declarative memory as separate entities) and is used to explicitly store memories with their associated features, represented as chunks symbolically with slots and their respective values in the declarative module. These memories are retrieved based on environmental cues for any given situation guided by sub-symbolic values \cite{anderson2007a}. The strength of association between current environmental cues and declarative memories enables retrieval. These cues are stored in buffers specific to the modules, and the communication between them is facilitated by the central procedural memory (production) system. 


While the benefit of systematically replicating human cognition can be applied to a wide variety of scenarios, the knowledge represented in ACT-R is mostly task-specific and is usually set manually by the modeler. This raises a serious bottleneck with respect to knowledge representation within the ACT-R architecture. \citeA{kelly_holographic_2020} have adapted holographic representations to replace the default declarative memory system of ACT-R, which they call holographic declarative memory (HDM). HDM uses distributional semantics to more store declarative memory chunks within ACT-R declarative memory in a more scalable way.
\subsection{Representing Racialized Worlds} 

 Individuals act not only on immediate evidence nor solely on personal prejudice but in context of “cultural worlds that afford or promote particular racialized ways of processing and seeing the world” \cite{salter_racism_2018}. People reproduce their worldview through “selected preferences”. \citeA{kelly_holographic_2020} models cognitive interpretation of textual statements as vector-symbolic semantic spaces.
 
 Such vector-based semantic spaces are useful as associations between concepts (e.g., through representations of similarities) can otherwise be set as association strengths between chunks in ACT-R’s declarative memory. Thus, having racial schemata (e.g., \citeNP{ray_racialorgs_2019}) represented in semantic space would present an opportunity for process-oriented understanding of impacts of such schemas on behavior. If one can effectively map the interaction between the semantic systems of racial schemata and the chunks of memory used for immediate decision making, one can begin to model how sociocultural worldviews create selected preferences.
 
\subsection{Vector-Symbolic Architectures} 

Vector symbolic architectures, or vector symbolic algebras, (VSAs) seek to combine many of the advantages of connectionism and symbolicism\footnote{For a more complete definition and discussion, refer to \cite{kleyko_survey_2022}.}. One such VSA is Holographic Reduced Representation \cite{plate_holographic_1995}, which has the advantage over other VSAs like the tensor product representation in having the binding of two vector "symbols" producing a vector of the same size. VSAs have been used for modeling hierarchical representation of concepts in semantic memory (\citeNP{eliasmith_how_2013, blouw_concepts_2016}). Holographic Declarative Memory implements ACT-R's Declarative Memory and learns representations of concepts from text tokens \cite{kelly_holographic_2020}. However, learning a hierarchical semantic representation from text that is directly compatible with structured, symbolic ACT-R declarative memory queries remains an open problem.

There is evidence that the brain's default mode network (DMN), which is activated during conceptual reflection, retrieves structured knowledge and constructs thoughts within a specific reference frame \cite{frankland_concepts_2020}. The representations seem to be amodal (as in not specific to written language, audio, or visual). Distinct sentence-like and map-like\footnote{We use "map" here in the sense of a function that maps from one continuous space to another} representations of concepts that are both stored and are complementary to each other. Sentence-like representations, which come from the language of thought (LoT), require the ability to create new concepts from old ones with structured rules, like a formally defined language. Map-like representations are needed when complex concepts often cannot just be generated by structured grammatically. Combining two words can require traversals of a conceptual space to properly represent them; merely assigning roles and binding in the same ways for any two concepts is not sufficient. In fact, there can be both intersective (within-object binding) and collective (across-object collecting) conjunctions. To put it in VSA terms, some concepts require binding between sub-concepts while other require bundling of higher-up concepts. 

However, despite the context needed for combining, there is evidence from BOLD data for temporal reuse of representations. Therefore, sentence-like and map-like representations must play a role. A good foundation for ensuring the two representations stay distinct is to enforce orthogonality between role, the type of structured argument a concept serves as, and filler, the meaning of the concept itself. Another important concept is that there must be fast probabilistic inference of “highly abstract prior knowledge” as opposed to traversing the conceptual space or a list of episodes to manually construct a thought every time there is a memory retrieval request (\citeNP{frankland_concepts_2020, kumar_semantic_2021}.).

\begin{figure}[H]
\begin{center}
\includegraphics[scale=0.1]{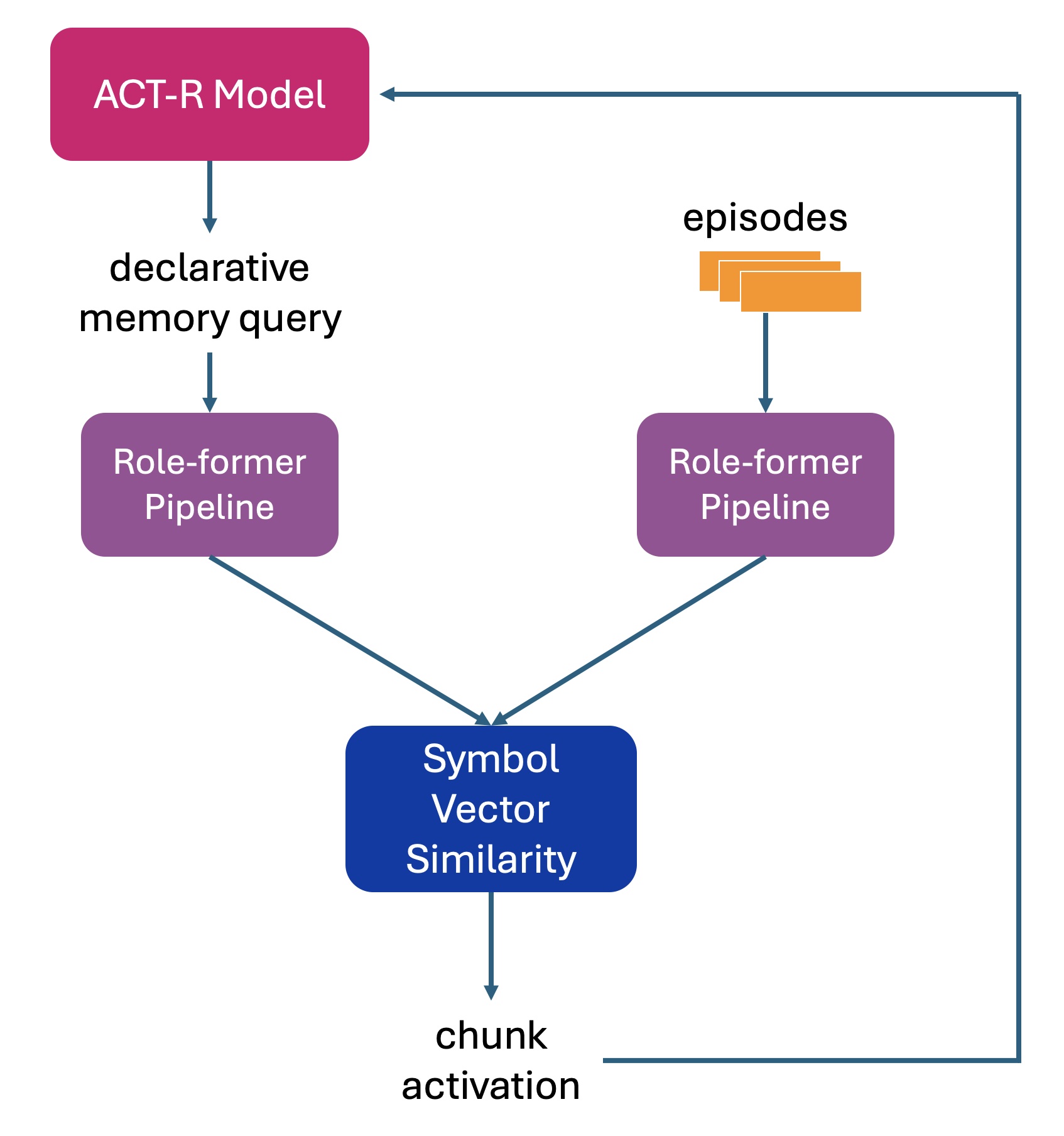}
\end{center}
\caption{Overview of our Model} 
\label{overview}
\end{figure}

Here, we propose an auto encoder model that keeps internal representations of roles and fillers with pseudo-orthogonal vectors and creates multi-level episode vectors that can allow for different levels of interaction between vectors, shown in Figure \ref{overview}.



\section{Methods}

\subsection{Autoencoding Roles}
\begin{figure}[H]
\begin{center}
\includegraphics[scale=0.1]{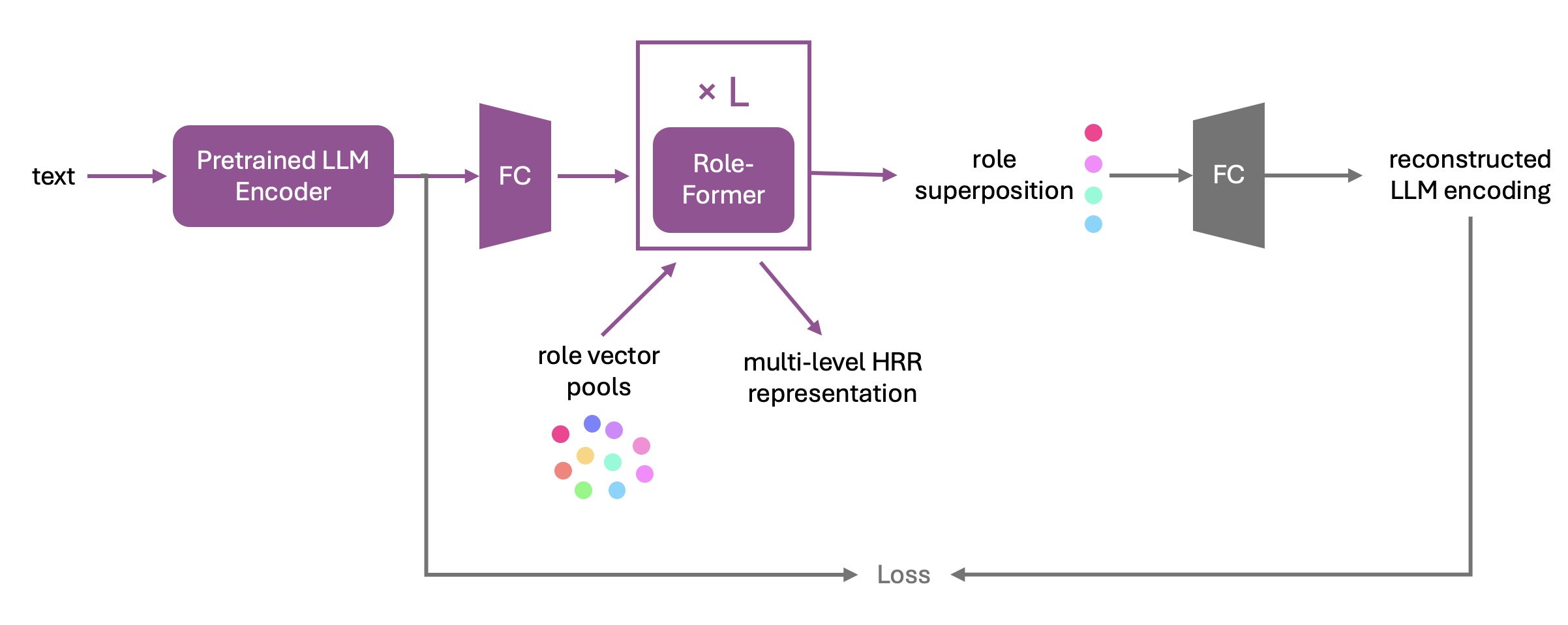}
\end{center}
\caption{Role-former pipeline. The parts highlighted purple are part of the pipeline in Figures 1 and 3, while the greyed out parts are only used during training. The Role-Former Block, shown in Figure \ref{role-former-block}, is repeated L times. FC here means fully connected linear layer. 
} 
\label{role-former-pipeline}
\end{figure}

\newcommand{\rit}{$R_i^{t}$}
Given a sequence of tokens $i = 1 ... n$, we pass it to an the BERT encoder and get an encoding $Enc_i \in \mathbb{R}^{d}$ where $d$ is the encoding dimension. We define \textit{role pools}, $ROLES_{i}$, unitary vectors initialized as in \cite{ganesan_learning_2021} for layer $i \in 1 ... L$. These vectors are the same regardless of network inputs -- we hypothesize that, given enough training examples, the auto encoder will pick certain vectors to fit certain roles. These "roles" are not limited to representing grammatical or semantic roles with other tokens in a sentence -- they could also represent ontological information and typical attributes of particular concepts. At the end of the role-former blocks, which will be explained next, the output is compared against the input to see how much of the original information was preserved (see \ref{loss-fn}). The overall autoencoder pipeline is shown in Figure \ref{role-former-pipeline}.

We pass each per-token LLM output into a fully connected linear layer with weight matrix $W \in \mathbb{R}^{d \times |ROLES_{i,j}|}$. To each $|ROLES_{i,j}|$-length vector, we apply the softmax to get a weighing of each vector in each pool which we call $R_i^{t}$. We take the dot product between \rit and $ROLES_{i,j}$, treating the pool as a set with a predefined ordering into a matrix, to get the role superposition $RS_{i,j}^1$. In the final loss function, which will be explained at the end of this section, there will be a term that penalizes too wide of a spread in \rit, in order to ensure each layer meaningfully "picks" a small set of roles. However, we do not simply take the single role vector corresponding to the maximum of \rit in order to allow for nuances of meaning; for example, a word can be the object of a verb in one case and the subject of a verb in another case.

As we go layer by layer, we may want to maintain a multi-level HRR representation of each token that can be easily separated by layer later on. There is evidence that multiple levels of representations are needed for "slow" and "fast" retrieval and reasoning (\citeNP{frankland_concepts_2020,kumar_semantic_2021,eliasmith_how_2013}). The HRR the network creates at each step may not be directly suitable for this task due to the ways roles are combined and earlier layers decay as one goes deeper within the network. Therefore, for purposes in which we need to save the all layers of representation for a particular input, we maintain an accumulated output HRR $Multi_i, i \in 1 ... n$ that will be used in a later step. This $Multi$ vector is not used during the auto-encoder training since we mainly optimize loss using the final layer output, not intermediate layer activations. However, but we do save $Multi$ when processing queries and episodes so we can calculate query-episode concept similarities to estimate chunk activation.

\begin{figure}[H]
\begin{center}
\includegraphics[scale=0.1]{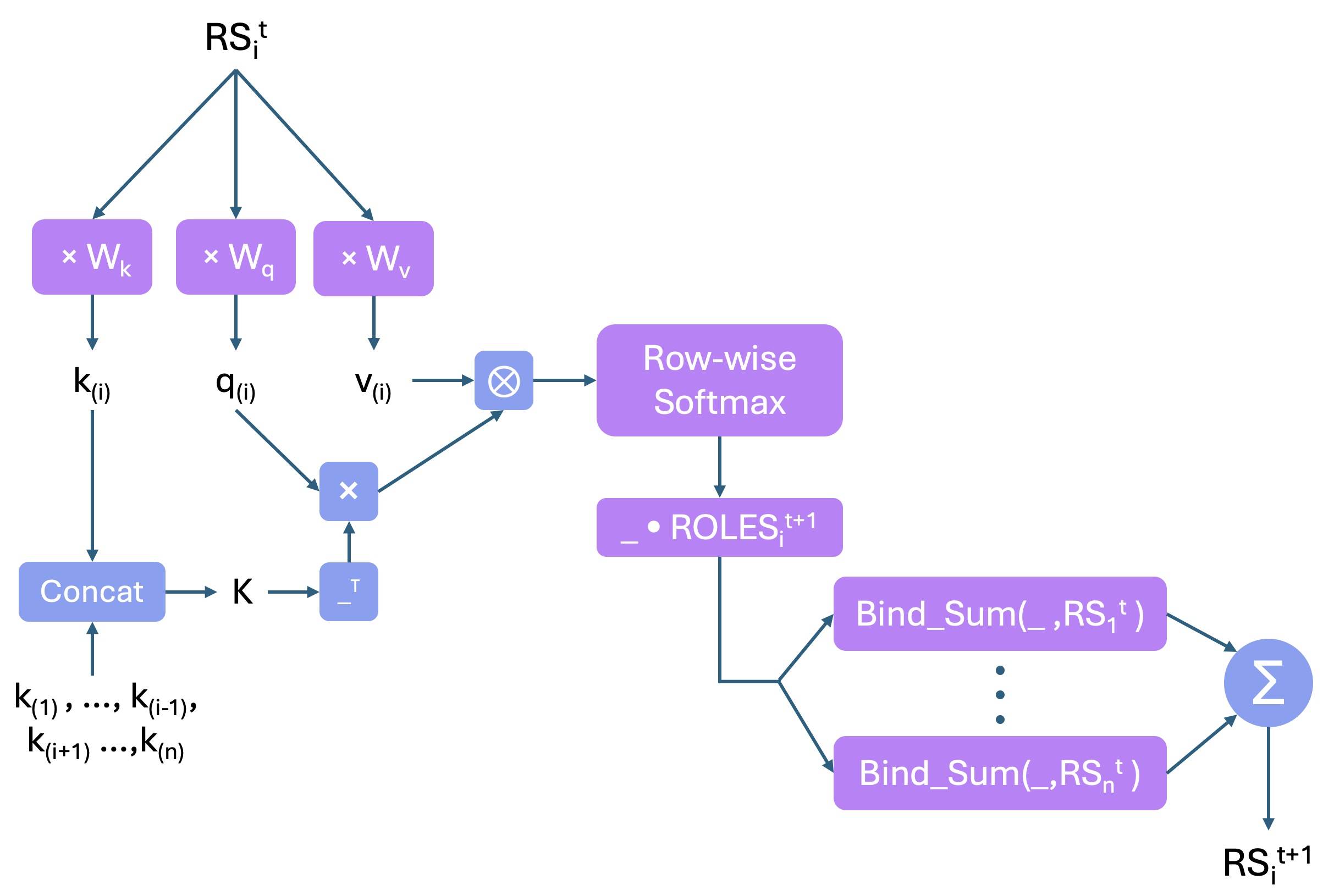}
\end{center}
\caption{Role-former. We show the processing of one token at index $i$.The block takes $i$'s role superposition vector from the $t$'th layer, $RS_i^t$, and outputs the token's role superposition vector for the next layer, $RS_i^{t+1}$. Notation: $\times$ is matrix multiplication, $\otimes$ is outer product, $\cdot$ is dot product, and $\_^T$ is transpose.} 
\label{role-former-block}
\end{figure}

We sum the $RS_{i,j}^1$ and get a single role superposition $RS_i$. This is the first level of output that we add to $Multi$:
\begin{equation}\label{multieq}
    Multi \longleftarrow Multi +  RS_i \odot \mathop{SYM}_i \odot L^1
\end{equation}
where $\mathop{SYM}_i$ is a fixed HRR vector that maps to the token $i$.
$L$ is a fixed HRR referring to the layer and we raise $L^i$ for layer index $i=1$. Here, and from now on, we use $\odot$ to mean the binding operation \cite{plate_holographic_1995}. We now pass $RS_i$ into the first "role-former" block, depicted in Figure \ref{role-former-block}.

\newcommand{\rsit}[0]{RS_i^t}

The role-former receives an input $RS_i^t$, representing the $i$th token from previous layer $t$. 
We define keys, queries, and values as in a standard self-attention module \cite{vaswani_attention_2023}:
\begin{align}
k^{(i)} &= \rsit W_k \\
v^{(i)} &= \rsit W_v \\
q^{(i)} &= \rsit W_q \\
K &= concat(k^{(1)} ... k^{(n)})
\end{align}
where we have $W_k$ and $W_h$ are in $\mathbb{R}^{N \times H}$ and $W_v \in \mathbb{R}^{N \times |ROLES^{t+1}|}$ for hidden size $H$ and HRR dimension $N$. 

However, our goal differs from a normal transformer in that we are not satisfied with a single scalar weight being calculated per sequence - we want a vector of length $|ROLES^{t+1}|$ per sequence to know which role describes the relation between token $i$ and other tokens as well as with token $i$'s representation in the previous layer. We could have an attention head for each possible role, but that would impose a severe computational limit of how many roles we could have at each layer. Instead, we want to be able to calculate the $R_{i \xrightarrow{} \cdot}^{t+1}$, the role weighings between token $i$ and every token, as a matrix operation. So, we take the outer product rather than the usual inner product between $K^Tq^{(i)}$ and $v^{(i)}$:

\begin{equation}
    R_{i \xrightarrow{} \cdot} = \operatorname{softmax}(K^Tq^{(i)} \otimes v^{(i)} )
\end{equation}
where $\otimes$ is the outer product. Next, to go from weights to the actual relational role superposition, we take the dot product with the matrix of roles for this layer:
\begin{equation}
    RS_{i \xrightarrow{} \cdot} = R_{i \xrightarrow{} \cdot} \cdot ROLES^{t+1}
\end{equation}

However, we do not want to simply have new roles based on the old roles be the final representation. In order to have a shot at compositionality, we need to store the combination of the newly chosen relational roles and the existing roles to more explicitly indicate that the relationship is between each token's representation:

\begin{equation}
    RS_i^{(t+1)} = \sum_{i' = 1}^n \operatorname{BindSum} (RS_{i'}^t, RS_{i \xrightarrow{} i'}^t)
\end{equation}

where $\operatorname{BindSum}$ is defined as follows:
\begin{equation}
    \operatorname{BindSum(A, B) = \frac{1}{\alpha} (A \odot B) + \frac{1}{\beta}A + \frac{1}{\beta}B}
    \label{bindsum-eq}
\end{equation}
We do not simply use the binding operation on its own, because $A \odot B$ is entirely dissimilar from $B$ (aside from noise), but we do want to maintain similarity so representations with overlapping roles still share some similarity if they share any roles at all, but have a greater similarity still if they share \textit{combinations} of roles. The hyperparameters $\alpha$ and $\beta$ are there to control a blowup in the magnitude of the bundled HRR vector that could drown out the relative impact of newer role vectors being added. However, this does have the consequence of the impact of earlier layers diminishing exponentially, which could be a big problem if $\alpha$ and $\beta$ are too high. We account for this with the previously mentioned $Multi$ HRR vector, which contains equally weighted representations across layers by construction. We add $RS_i^{(t+1)} \odot \mathop{SYM}_i \odot L^{t+1}$ to $Multi$ at this role-former layer, as we did in Equation \eqref{multieq}.
Given that we are now on layer $t+1$, the output from the role-former is fed into the next role-former, until we reach the end. Next, the output is passed into a fully-connected layer of size $\mathbb{R}^{N \times (h*d)}$ to output a reconstruction of the original LLM encoding $Enc_i '$.

We train the network using a loss function that combines the standard reconstruction loss -- mean squared error between $Enc_i$ and $Enc_i '$ -- with the average entropy of the role weighting vectors \rit:

\begin{equation}
    \mathcal{L} = \frac{1}{nS}\sum_{i=1}^S \sum_{j=1}^n  \left[\Lambda||Enc_j - Enc_j' ||^2  + \lambda \sum_{t=1}^LH(R_{i,j}^t)\right]
\end{equation}\label{loss-fn}
for hyperparameters $\lambda$ and $\Lambda$ to weigh the two losses, batch size $S$, sequence length $n$, and number of layers $L$ . H(x) is the normalized information entropy function \cite{shannon_mathematical_1948}:
\begin{equation}
    H(R_i^t) = -\frac{1}{\log |ROLES_l|} \sum_{t=1}^{|ROLES_l|}R_t^t[j] \log(R[j])
\end{equation}
The function measures, intuitively, the uncertainty of selecting the given role for a token $i$ at layer $t$, divided by the maximum such uncertainty (the case where every option being has uniform probability).

\subsection{Autobiographical Memories and Episodes} \label{epis-section}
The cultural theorist Sylvia Wynter wrote of the “mechanisms of the brain that, when activated by the semantic system of each such principle/statement, lead to ... [how we] ... experience ourselves as this or that genre/mode of being human.” \cite{wynter_unsettling_2003}. While semantic memory is important for understanding social meanings, an account of self representations is important for the role racialization plays in an individual's actions. We argue that a cognitive architecture should also include \textit{autobiographical} memories to help provide such representations of self through architectural mechanisms. Autobiographical memories are, in part, made of episodic memories and life time period descriptions of individuals. In our modeling, we choose to store them separately from semantic (declarative) memory while also modeling their interaction.

Though we do not have corresponding autobiographical data for participants in available IAT datasets (which could be used for episodic memory), we approximate the impact of episodic memory using other datasets. We use oral history projects that ask people about their lives and document the race of interviewees. For example, the Black Women Oral History Project archive \cite{letitia_woods_brown_interviews_2013} contains transcripts of over sixty interviews with Black women who "had made significant contributions of varying kinds to American society in the early and mid-decades of the twentieth century." Importantly, the interviewers specifically asked them about the "ways in which being Black and a woman had affected their options and the choices made" throughout their life. 

For each transcript, we manually selected the 10 page portion most relevant to formative experiences in childhood and early adulthood. Then, we ran the PDFs through OCR and text extraction. Both steps likely did introduce some error; upon visual inspection, the majority of lines did seem to be processed correctly however. The only preprocessing step was to remove empty lines.

We account for self representations using these datasets by manually coding words in transcripts that refer to the self ("I", "me", "my") with the original symbol vector added to the $\mathop{SYM}_{BLACK}$ for use in the IAT task. In the future, we could use reference tracing to dynamically make these assignments or even to allow for gradations in self-ness. The episode text, along with corresponding per-token symbols, was passed to the already-trained role-former pipeline shown in Figure \ref{role-former-pipeline} in order to give us per-token HRR $Multi$ vectors (see equation ). These HRRs were then summed together to give a single per-episode HRR vector. 

\subsection{ACT-R Model for IAT} 

We use the implicit association task (IAT; \citeNP{greenwald1998a}) to test our model. IAT uses latency-based measures such as response time to investigate automatic association with certain themes. The task we chose is a race-based implicit association task, in which pleasant or non-pleasant attributes are associated with images of Black or White people. In this task, participants are shown stimuli in 5 blocks. The first block presents stimuli related to pleasant and non-pleasant attributes, and participants are expected to press the associated key. Words used for pleasant stimuli are joy, pleasure, happy, love, good, and words used for non-pleasant stimuli are bad, agony, evil, hurt, and nasty. In the second block, the target associations are displayed, which, in our case, are randomized pictures of White and Black individuals. In the third block, which in our case is the congruent condition, the pictures of White individuals were mapped to the same key as that of pleasant stimuli. For the next round, which acts as training for the incongruent condition, the key mapping for Black and White stimuli is switched to map to the opposite keys. In the final round, which tested mapping under incongruent conditions, pictures of Black individuals were mapped to the key corresponding to pleasant stimuli, and vice versa. 

 We collected the data from 120 Black/African American and White/Caucasian participants, cleaned this dataset to remove data from participants whose response times were greater than 10000 ms or less than 300 ms. This left us with data from 104 participants. We compare the response times by blocks with that of the model we developed for this task, averaging out for all participants. These response times were scaled to account for the penalty for wrong key presses. We encoded the words displayed to user in ACT-R's DM and modified base-level parameters and used set-similarities to adjust the similarities between chunks to obtain the fit.  



\subsection{Training the Roleformer}
We trained a roleformer auto encoder (RAE) with two layers of roleformers. We set the $\alpha = 10/4$ and $\beta=10/3$ in the equation \ref{bindsum-eq}. The pools were set to 91, which is about the minimum $n$ such that ${n\choose3}>117000$, the number of synsets in WordNet \cite{princeton_university_21_2010}. We don't have a penalty for entropy on the first role pool, so the network was free to pick amongst the 91 HRR vectors. We did this to allow for the nuances of per-word semantic categories that could then be operated on more simply by roles in later layers. The second pool size is 56, the number of semantic macro-roles a noun, verb, adjective, or adverb was proposed to play by \cite{bentley_cambridge_2023}. Finally, the last pool has 36 vectors, for the number of roles in ConceptNet \cite{speer_conceptnet_2018}. We hypothesized once the per-word semantic and relational tokens were represented in the first two layers, the third layer would represent the less context-specific, abstract relationships between concepts. The entropy loss weight was 30\%, with the mean squared error weight left at 1. The HRRs were of size 1024 and the hidden layer dimensions for the roleformer blocks were 100 and 75 for the two layers. In total, this gave us a model with 1,311,765 32-bit float parameters. 

We used a dataset of ten interview transcripts from \cite{letitia_woods_brown_interviews_2013}. These are the same source from which the episodes were collected; we also considered training on another dataset (e.g. Wikipedia text), but it made sense to train the model within the same domain since oral history texts would very likely use a different distribution of language than Wikipedia. We processed the text in the same way as the episodes in \ref{epis-section} but without adding $\mathop{SYM}_{BLACK}$ to self-referring tokens. We arrived at a dataset with 5,516 lines of text. During training, these examples were shuffled to prevent order-dependent effects during training. This was necessary, but likely did erase some ordering information. We split the lines into 80\% train, 10\% validation, and 10\% test sets so we could prevent overfitting and study generalization performance.

We trained for 68 epochs with a learning rate of 0.001 and the AdamW optimizer, stopping when the model began to overfit. The RAE achieved a 0.0164 weighted reconstruction loss on the training set, 0.0175 on validation, and 0.0224 on the test set. 


\section{Results}

\subsection{Reproducing The Conjunction Fallacy}
The conjunction fallacy is a documented phenomenon in which humans sometimes judge the probability of two events occurring together as higher than the probability of each event individually, which violates classic probability rules \cite{tversky_extensional_1983}. This may be because combining two concepts creates a new meaning in a semantic vector space, and this meaning can be compared to another with cosine similarity \cite{kelly_holographic_2020}. We used the second cosine similarity test as \citeA{kelly_holographic_2020} to compare the memory vectors in conjunction versus separately: $\operatorname{cos}(rs_{linda}, rs_{feminist}+rs_{bankteller}) > \operatorname{cos}(rs_{linda}, rs_{bankteller})$ for $rs_{tok}$ as the final output role superposition of the role auto encoder pipeline for the token $tok$. However, unlike \cite{kelly_holographic_2020}, we did not construct Linda as a sum of memory vectors but rather used the vector for Linda corresponding to the token after passing the description from \citeA{tversky_extensional_1983} into the role-former pipeline. We found that the first similarity was 0.022\% greater (absolute difference) than the latter. When we replaced $linda$ with $kathleen$, the latter of which did occur in our training set, the difference -- 0.018\% -- was slightly less. When we considered the $Multi$ vectors rather than the role superpositions, the difference was the largest at 0.68\%.
A relatively small training set and a lack of penalty term in the loss function for grouping items close together may explain why the differences in similarities are small, even if the semantic vectors do exhibit the fallacy.

\subsection{Fitting to the IAT}
We calculated reaction times on two different IAT blocks using both per-token and episodic HRR outputs. Per-token vectors were the direct result of passing the single word e.g. "black" through the role auto encoder (RAE) pipeline. Episodic vectors were obtained by passing the episode texts described in \ref{epis-section} into the RAE and then convolving with the inverse of $L^1$ and the inverse of $SYM_{black}$ to get the summed roles for $SYM_{black}$ in the episodes. 

In Round 3, participants had to identify White stimuli as pleasant and Black stimuli as unpleasant. We can represent this as a similarity between a retrieval query and a chunk: the cosine similarities between "white" and "pleasant" and between  "black" and "unpleasant", which we average together. For per-token representation, the averaged similarity was 0.997348 while for the episode-level representation, the averaged similarity was -0.0378. In the latter case, this indicates the concepts were orthogonal but leaning slightly towards opposites. 
In Round 5, participants needed to identify Black stimuli as pleasant and White stimuli as unpleasant. For per-token, the averaged similarity is 0.996907 while for episodes the averaged similarity is 0.1794.

We used the formula for expressing an ACT-R activation from \citeA{kelly_holographic_2020}: 
\[
    A = \ln(\frac{C^2}{1-C^2})
\]
where C is the cosine similarity between the cue and the chunk. Next, we pass the activation to the ACT-R retrieval time formula \cite{anderson2007a}
\[
    RT = Fe^{-fA}
\]
for adjustable parameters latency factor value $F$ and latency exponent value $f$. We set $F=1.219$ / $F=0.9$ and $f=0.02$ / $f=0.025$ for the single-word and episode vector similarities respectively. For the Black participants group, the episode vector, and the ACT-R model fit to all participants, the Round 3 RT was higher than the Round 5 RT (see Table \ref{IAT-RT-table}).  

\begin{table}[H]
\begin{center} 
\caption{IAT Reaction Times Fit} 
\label{IAT-RT-table} 
\vskip 0.12in
\begin{tabular}{lll} 
\hline
Agent   & Round 3 & Round 5 \\
\hline
Black Participants & 1051  & 960\\
White participants   &  1093 & 1100 \\
Overall Participants  &   1073  & 1035   \\
ACT-R Model       & 1044  & 964 \\
RAE Single-Word Vecs & 1098 &  1101\\
RAE Episode Vecs & 1060 & 980 \\
\hline
\end{tabular} 
\end{center} 
\end{table}

\section{Discussion and Conclusion}
Our selection of an LLM as the first input source for memory encoding was intentional -- they encode a wide array of racial beliefs and structures. The better fit of the LLM-trained role auto-encoder model, without modification, to the White participant IAT RTs rather than Black participant IAT RTs, may lend support to the theory that LLMs encode the worldviews of dominant social groups (\citeNP{dancy_cogarch-genai_2023, dancy_ai_2022}). However, when we used the roles for "Black" from the RAE-encoded episode vectors, the effect was reversed. Accounting for the role of the self with episodic memory vectors, in conjunction with separating form from content with the role autoencoder, may allow us to represent the perspective of a Black self better than using an LLM-based memory encoding alone.

\subsection{Limitations / Future Work}
As a practical data limitation, we do not have autobiographical accounts corresponding to specific IAT particpants - we only do racial demographic matching. This could be a worthy area for future study.

The bioplausiblity of standard backpropagation, our training method for the role-former pipeline, is debatable \cite{lillicrap_backpropagation_2020}. There is some evidence for the bioplausibility of certain fast weight programmer learning rules \cite{irie_fast_2025} This is a promising avenue to explore in the future. 

In this paper, we treat episodes as static. Episodic memories in reality change over time in part due to memory retrieval requests. However, the vector-symbolic structure of our episode vectors would allow episode vectors to be updated by simply adding context in the way memory vectors are accumulated in \cite{kelly_holographic_2020}, by weighing episode vectors by recent similarity like how declarative memory chunks are weighted in ACT-R \cite{anderson2007a}, or by using fast weight programmers \cite{irie_fast_2025} to adjust the episode vectors per retrieval request. 

The inputs we considered to both the FWP and Role-Former were both text, or more broadly a sequence of discrete tokens. Continuous data would need to be adapted in order to fit into this framework. This could be done via splitting images into  patches, the way image transformers work, or using the attention heads of a larger visual model.

Unlike normal transformers, the role-former did not use a positional encoding; our rationale is that the LLM output already contained positional information. However, we could explore using positional encodings. For example, could use a predefined $POS$ symbol raised to position index $i$. Relative positional encodings may be more flexible, and smoother positional encodings such as \cite{ray_adapting_2025} and \cite{vaswani_attention_2023} exist, but we are unaware if these functions are still smooth in the HRR similarity space. 

We used a model based on a relatively old LLM, BERT, that does not exhibit full training data transparency \cite{devlin_bert_2019}. Newer research-oriented LLMs do better on this \cite{workshop_bloom_2023}, but they are mostly decoder-only architecures, which are unsuited for our purpose since they are created to generate text rather than to be a representation for a non-strictly-generative downstream task. There is a proposed method to get encodings from decoder-only architectures \cite{qiao_decoder-only_2025}, which we may want to use later instead of the BERT-based model .

Finally, we specifically chose the role-former architecture such that the final output of the role-former encoder pipeline is in the set enclosed under an HRR vector distribution and the accompanying bundling and binding functions. A resonator network could be used to factor output for model interpretation studies, obviating the need for cleanup memory when decoding large sequences \cite{frady_resonator_2020}.

\section{Acknowledgments}

This material is based upon work supported by the AI Research Institutes Program funded by the National Science Foundation under the AI Institute for Societal Decision Making (NSF AI-SDM), Award No. 2229881. 

This work was supported by the National Science Foundation under grant No. 2144887. 



\bibliographystyle{apacite}

\setlength{\bibleftmargin}{.125in}
\setlength{\bibindent}{-\bibleftmargin}

\bibliography{latex/refs}

\end{document}